\documentclass[10pt,twocolumn,letterpaper]{article}

\usepackage{cvpr}              %

\usepackage{graphicx}
\usepackage{amsmath}
\usepackage{amssymb}
\usepackage{booktabs}
\usepackage{multirow}

\graphicspath{ {./images/} }
\newcommand{\para}[1]{\vspace{0.3em} \noindent \textbf{#1} \hspace{0.5em}}
\newcommand{\paranospace}[1]{\noindent \textbf{#1} \hspace{0.5em}}

\definecolor{cvprblue}{rgb}{0.21,0.49,0.74}
\usepackage[pagebackref,breaklinks,colorlinks,allcolors=cvprblue]{hyperref}

\def\paperID{6582} %
\def\confName{CVPR}
\def\confYear{2025}

\title{If At First You Don't Succeed: \\ Test-time Re-ranking for Zero-shot, Cross-domain Retrieval}

\author{Finlay G. C. Hudson\\
Department of Computer Science\\
University of York\\
York, United Kingdom\\
{\tt\small finlay.gc.hudson@york.ac.uk }
\and
William A. P. Smith\\
Department of Computer Science\\
University of York\\
York, United Kingdom\\
{\tt\small william.smith@york.ac.uk}
}

\begin{document}
\maketitle
\begin{abstract}
In this paper, we introduce a novel method for zero-shot, cross-domain image retrieval. Our key contribution is a test-time Iterative Cluster-free Re-ranking process that leverages gallery-gallery feature information to establish semantic links between query and gallery images. This enables the retrieval of relevant images even when they do not exhibit similar visual features but share underlying semantic concepts. This can be combined with any pre-existing cross-domain feature extraction backbone to improve retrieval performance. However, when combined with a carefully chosen Vision Transformer backbone and combination of zero-shot retrieval losses, our approach yields state-of-the-art results on the Sketchy, TU-Berlin and QuickDraw sketch-based retrieval benchmarks. We show that our re-ranking also improves performance with other backbones and outperforms other re-ranking methods applied with our backbone. Importantly, unlike many previous methods, none of the components in our approach are engineered specifically towards the sketch-based image retrieval task - it can be generally applied to any cross-domain, zero-shot retrieval task. We therefore also present new results on zero-shot cartoon-to-photo and art-to-product retrieval using the Office-Home dataset.
Project page: finlay-hudson.github.io/icfrr, code available at: github.com/finlay-hudson/ICFRR
\end{abstract}
    
\section{Introduction}

The problem of zero-shot, cross-domain retrieval involves matching a query image of an unseen class against a set of gallery images of a different domain to the query. Common domains of interest include photos \cite{liu2021image}, sketches \cite{eitz2010sketch}, cartoons \cite{talib2013weighted}, clip art \cite{hu2022feature}, schematics \cite{yang2020diagram} and maps \cite{lu2021cross}. Retrieval amounts to ranking the gallery images against the query according to some similarity measure \cite{yelamarthi2018zero}. This is particularly challenging in the zero-shot, cross-domain setting since correctly matched query-gallery pairs may contain no shared visual features. Instead, they may only match in an abstract semantic sense but, since the class is unseen, the model may not have learnt appropriate semantic concepts that enable matching \cite{zhang2017learning}. Due to the inherent dissimilarities in visual features between query and gallery images, the initial ranking will inevitably comprise of a mixture of both correct and incorrect matches. This disparity arises from variations such as missing elements in one domain, perspective changes, scale differences, and more. To address this challenge, we exploit the visual feature similarities among gallery-gallery image pairs, enabling the creation of a semantic chain of meaningful matches, facilitating the elevation of same-domain matches within cross-domain rankings. This approach emulates the cognitive mechanism employed by the human brain, leveraging thematic semantic systems to establish meaningful connections between concepts \cite{mirman2017taxonomic}.

\begin{figure}[!t]
\centering
\includegraphics[width=1.0\columnwidth]{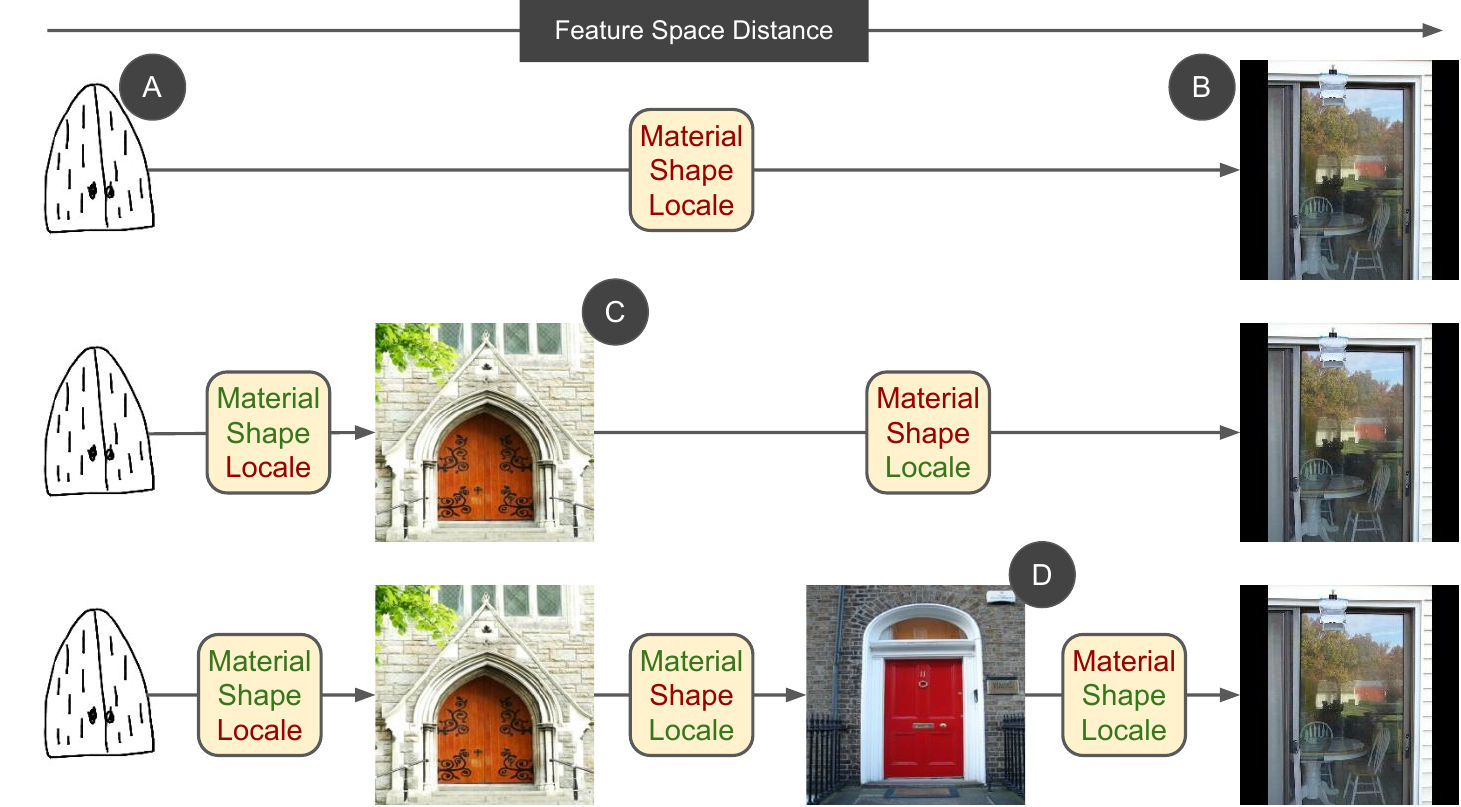}
\caption{Re-ranking rationale: (A) is a sketch of an ornate wooden door while (B) is a modern glass sliding door. These images share almost no visual similarities, only semantic similarities. Re-ranking at test-time can help bridge the visual gap by exploiting relationships between same-domain gallery images (C, D).}\label{fig:reranking rationale}
\end{figure}

These semantic links enable the identification of initially poorly-ranked gallery images that are actually conceptually connected to the query. As a result, these initially poorly-ranked gallery images are rewarded and gradually moved up the ranked list, making room for the discovery of more distantly-connected gallery images through multiple gallery-gallery matches. Through iterative application of this process, our approach enables the exploration and identification of gallery images that exhibit stronger semantic connections to the query, even when the initial ranking might not capture these associations. This iterative refinement process facilitates the discovery of relevant images that may have been overlooked in the initial ranking, ultimately enhancing the overall accuracy and effectiveness of our zero-shot, cross-domain image retrieval method. Our re-ranking process uses ranks as opposed to absolute distance measures in some embedding space. This provides robustness and avoids making any assumption about the distribution of distance or the need for any normalisation.

We provide a concrete example in Figure \ref{fig:reranking rationale} for a query sketch against gallery photos, though note that our approach is not limited to sketch-photo retrieval. The query sketch of some wooden doors (A) initially assigns a low rank to the gallery glass doors (B), since they do not match in material (wood versus glass), shape (curved versus square) or locale (without and with door frame). However, a gallery image of church doors (C) does share some visual features with the sketch (curved door shape, wooden material). The church doors contain some similarity to the glass doors (both are within door frames) however the shape and materials are different and so (B) is ranked relatively low in the gallery-gallery ranks for (C). However, the gallery image of the street house doors (D) are similar enough to (C) to be highly ranked (both contain a wooden door within a door frame) and, in turn, the house doors (D) are similar enough to the glass doors (B) to be highly ranked in the gallery-gallery ranks for (D) (rectangular door, within door frame). Our iterative re-ranking process is able to discover such semantic links and raise the rank position of difficult matches such as (B) against a very different query (A).

We combine this contribution with carefully chosen architecture design decisions, incorporating elements such as the Vision Transformer architecture \cite{dosovitskiyimage}, the Hard Mining Cross Domain Triplet Loss \cite{huang2021modality}, and a PK sampling Dataloader strategy \cite{zhai2019defense} to achieve state-of-the-art performance on the TU-Berlin Extended \cite{eitz2012humans}, Sketchy Extended \cite{sangkloy2016sketchy, liu2017deep} and QuickDraw Extended \cite{dey2019doodle} datasets. Furthermore, we demonstrate the versatility of our method by extending its application beyond sketch-based retrieval, showcasing its effectiveness on the Office-Home \cite{venkateswara2017deep} dataset for cartoon- and art-based retrieval.

\section{Related Work}

\paranospace{Re-ranking}
Sketch-based image retrieval (SBIR) is, in essence, a ranking problem. A model's success is determined by its ability to rank images of the same class as a query sketch higher than images from other classes \cite{sain2021stylemeup}. However, sometimes a sketch does not share direct similarities with all real-world images of the same class, presented in Figure \ref{fig:reranking rationale}. To attempt to alleviate this, re-ranking methods are employed to allow for these more distant connections to be exploited. \cite{wang2019sketch,wang2019enhancing} utilise a clustering re-ranking method with the underlying rationale that if a model assigns high rankings to multiple images from the same cluster, all images within the cluster should be ranked highly. However, clustering techniques can lack robustness \cite{hassan2019clustering} and in a cross-domain setting, a cluster centre of the same class may be far from a query domain image in visual feature space yet still be semantically related. In this paper we present a novel re-ranking approach that does not rely on clustering. Our method is more closely linked to the method proposed in \cite{zhong2017re}, a re-ranking method which involves combining original distances with IoU-based distances on k-reciprocal nearest neighbour sets. However, unlike our approach, their method lacks the incorporation of rank information in the re-ranking process and is also non-iterative, making it less effective in addressing scenarios where two distant elements in the initial list are indirectly connected through an intermediary element. \cite{ouyang2021contextual} perform the re-ranking task by refining affinity features, based on inner products between features for query and anchor images, with a transformer encoder. Contrary to our approach, this method exploits context between query images whereas our re-ranking method using a single query image, independent of query image distribution. Furthermore, it necessitates re-training for substantial different datasets or different input embedding dimensions, lacking the ``plug-and-play'' capability inherent in our method. 
Leveraging context from multiple query images is a common approach in various re-ranking methodologies \cite{zhang2023graph, zhong2017re, ouyang2021contextual}, with nearest neighbour graphs \cite{zhong2017re}, graph convolution networks \cite{zhang2023graph} or affinity features \cite{ouyang2021contextual} being utilised to combine information from multiple query images for query expansion. This make comparisons difficult for ZS-SBIR as the paradigm requires each query image to be considered individually, with no information leak. These methods are commonly used for Person Re-identification wherein the query itself is a track, therefore you have multiple instances of the same query, so it makes sense to share context between them. \cite{chum2007total, gordo2017end, iscen2017efficient, shao2023global} treats each query image individually, while \cite{sarfraz2018pose, ouyang2021contextual} allows for query-query information to be omitted in their implementation, hence we can compare against these methods.

\para{ZS-SBIR} 
\emph{Zero-shot} sketch-based image retrieval (ZS-SBIR) is a special case of SBIR in which the sketch and gallery image classes are never seen during train time \cite{tan2021survey}. The requirement of zero-shot ensures that the model cannot over-fit to the training set by learning very specific aspects of classes but instead has to learn the semantic understanding of how a sketch can relate to an image; making it more representative of a real-world use case \cite{tan2021survey}, whereby any category of image could be presented to the model. To help guide the model into learning a semantic relationship existing approaches attempt to use a joint representation space, to bridge the gap between the domains. \cite{gupta2022zero, dey2019doodle, liu2019semantic, dutta2019semantically, zhang2020zero} showcase the use of word embeddings to generate a semantic link between images but as  \cite{tursun2022efficient} suggests the use of word embeddings is not practical for many real-world use cases, with it being often difficult to describe a whole image by a single word or short phrase. \cite{gupta2022zero} use class names for text prompts as side information, this is however different to our paradigm as we work in a fully zero-shot manner, with knowledge of class names at inference time not being present. \cite{sheng2022sketch, zhan2022three, wang2019sketch, sain2022sketch3t, wang2023cross} generate edge-maps from the photographic images to imitate sketches while \cite{ren2023acnet} utilises a Cycle-GAN \cite{zhu2017unpaired} to convert the sketches to a photograph-like modality. The use of edge maps makes the solution task specific, if sketches were changed to cartoons this method would no longer work. %
Both edge maps and GANs also pose challenges in terms of adding noise to the system. Edge maps can result in noise due to sub-optimal edge selection, while GANs may generate inaccurate images due to limited representation learning. %

\para{Query Expansion and Diffusion}
The discovery of distant semantic connections between images is closely related with query expansion or diffusion techniques in retrieval. Query expansion \cite{qian2015enhancing, chum2007total, ahmed2020query, shiri2002thesaurus, banerjee2018relevance} involves initiating new queries based on related terms, features, or concepts related to the original query. Conversely, diffusion \cite{bai2018regularized, iscen2017efficient, donoser2013diffusion}, uses an offline-constructed neighbourhood graph to efficiently explore the data manifold during queries. Query expansion suffers from relying on hand-crafted features such as spatial information \cite{chum2007total}, colour and texture information \cite{ahmed2020query} or edge-maps \cite{qian2015enhancing}, making for a non-general solution that is hand engineered to specific problems, whereas our solution is applicable to any cross-domain problem. Diffusion assumes that distance in the feature space between nearest neighbours is consistent and comparable over the manifold which is often not the case \cite{liu2021handling} for cross-domain problems. %

In the most general sense, our re-ranking approach shares some similarities to a diffusion process. Traditional diffusion methods propagate information from one node to its neighbours over a graph based on some similarity measure. This is typically done using random walks, heat kernels, or graph neural networks. In our method, we are also propagating information (in this case, ranks turned into scores) iteratively across the gallery items. While this is not graph-based in the traditional sense, it is still a form of iterative information propagation. However, there are several fundamental differences. In our method, the gallery-gallery similarities (actually scores derived from ranks) never change. So, were our method to be viewed as diffusion, it could either be seen as a fully connected graph in which only query-gallery edge weights change or a star graph in which only query-gallery edges exist. The propagation of information is only via query-gallery similarities. Crucially, all of our updates depend only on rank information, converted to scores via a saturating function. This provides robustness to extreme differences in scores and avoids the need for hyperparameters related to the absolute scale of similarity values.

\section{Methodology}

Consider a training set of $N$ images $\mathcal{X}=\{(x_i^m,y_i)|y_i\in C_\text{train}, m\in\{A,B\} \}_{i=1}^N$ drawn from two different image domains, $A$ and $B$. Each training pair comprises an image $x_i^m\in\mathbb{R}^{H\times W\times 3}$ and a corresponding class label $y_i$ drawn from a set of possible classes $C_\text{train}$. Dissimilarity between images is defined by a distance function $D(x_i^{m_1},x_j^{m_2})=\|f_\theta(x_i^{m_1})-f_\theta(x_j^{m_2}) \|_2$ which measures Euclidean distance between embeddings of the images by an embedding function $f_\theta:\mathbb{R}^{H\times W\times 3}\rightarrow\mathbb{R}^{E}$ where $E$ is the dimensionality of the embedding, $\theta$ the parameters of $f$ and $\|f_\theta(x_i^m)\|_2=1$ (i.e.~embeddings are L2 normalised).

At test time, we are provided with a \emph{query} image, $x_Q^{A}$ of domain $A$ and a \emph{gallery} of $G$ images of domain $B$, $\mathcal{G}=\{x_i^{B}\}_{i=1}^G$. The goal is to solve this cross-domain retrieval problem by ranking the gallery images according to similarity to the query image. In the zero-shot case, the classes of both $x_Q^{A}$ and all elements of $\mathcal{G}$ are unknown but are drawn from $C_\text{test}$, with $C_\text{train}\cap C_\text{test}=\varnothing$ - i.e.~neither query nor gallery classes have been observed during training. Each query is treated independently and information cannot be shared between query images within a test set. Our objective is to learn the parameters $\theta$ such that the embedded features perform well for the task of zero-shot, cross-domain retrieval. This is a metric learning problem.

\begin{figure*}[!t]
  \centering
  \includegraphics[width=0.8\textwidth]{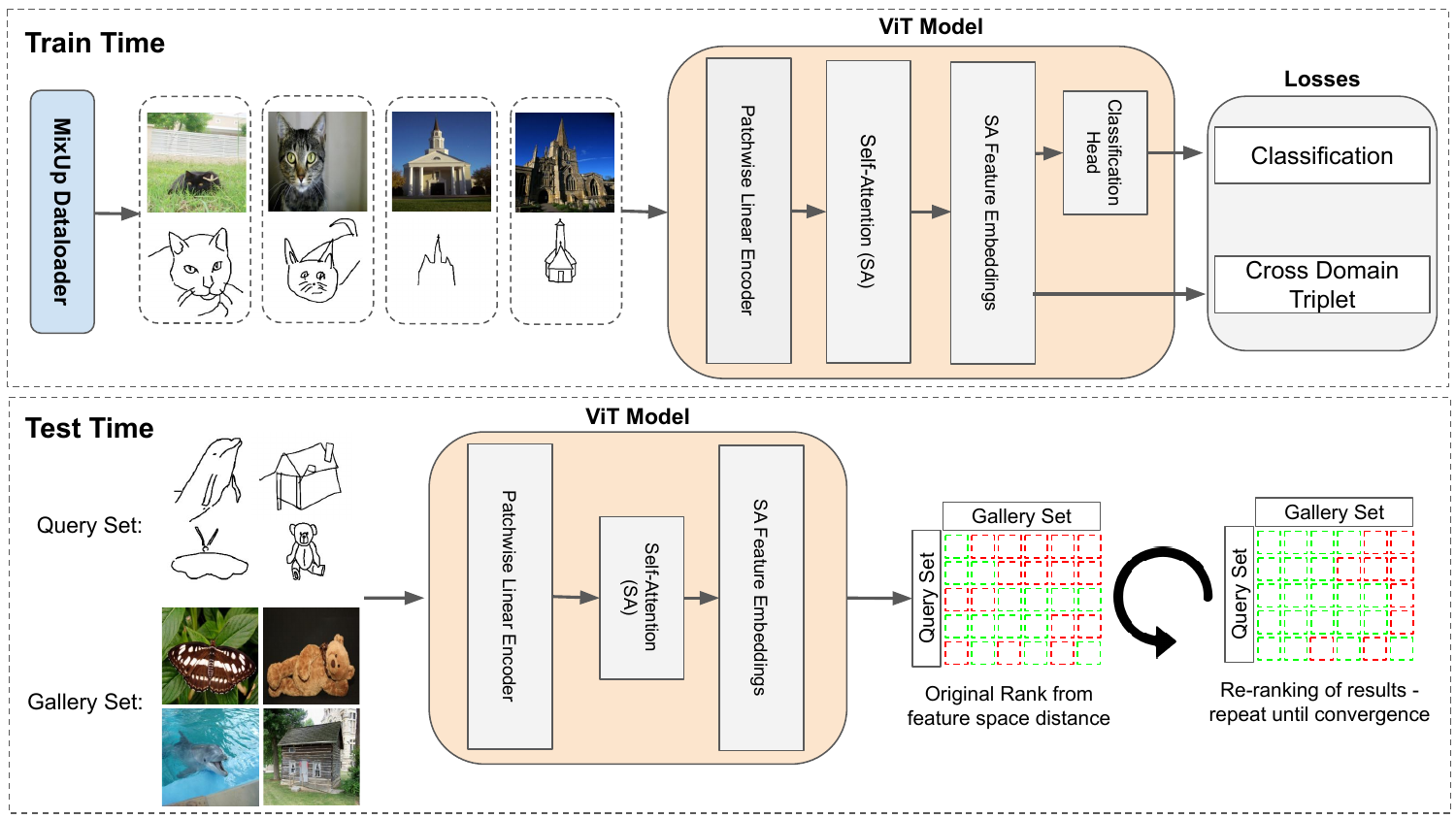}
\caption{
Overview of the architecture and operation of our approach. During training, image pairs are fed into a Vision Transformer model, producing embeddings and class predictions. These embeddings are used for both the Cross-domain Hard Example Triplet loss and the Cross Entropy loss. At test time, individual images are passed through a Vision Transformer to generate embeddings. Gallery image embeddings are then ranked based on their distances from a Query image embeddings, followed by re-ranking using our Iterative Cluster-free Re-ranking process.}
  \label{fig:architecture}
\end{figure*}

\subsection{ViT-x Architecture}
In practice, our embedding network is a Vision Transformer (ViT) \cite{dosovitskiyimage}. We chose to use a ViT base model with input size of 224x224 and 14x14 patches, along with a single class token. Our architecture is shown in Figure \ref{fig:architecture} for both train and test time modes of operation. During training time, we utilise \emph{cross entropy loss} and \emph{cross-domain hard example triplet loss}. We call the combination of this architecture, losses and training strategy ViT-cross (\emph{ViT-x}). During test time we introduce our novel \emph{Iterative Cluster-free Re-ranking} approach to leverage gallery set relationships to improve query-based gallery retrieval.

We train the ViT network to minimise a weighted sum of two losses:
$L_\text{total} = {\lambda_\text{Triplet}}L_\text{Triplet} + {\lambda_\text{CE}}L_\text{CE}$,
where the $\lambda$ are the weights, $L_\text{Triplet}$ is the cross-domain hard example triplet loss and $L_\text{CE}$ the cross entropy loss.
We now describe each of these losses in more detail.

\para{Cross-domain Hard Example Triplet Loss}
Triplet loss is commonly used within zero-shot tasks as it allows for metric learning of a feature space such that a cluster of elements from one class are far apart from the feature space representations of other classes. It is hoped that this structure is preserved even for unseen classes. The basic triplet loss \cite{schroff2015facenet} was extended by the MATHM method \cite{huang2021modality} to better apply to the cross-domain setting. The idea is to form triplets within domain A and within domain B but also across domains A and B. This ensures that the model is learning a feature space representation that clusters within-class and separates between-class for both domains. Another feature of this loss is that it selects the hardest examples per iteration batch. This is to ensure the model is constantly learning and does not always get triplet sets that it has already managed to easily separate.

Given anchor (anc), positive class match (pos) and negative class match (neg) images from both domains, we define the triplet loss in terms of the domains of the anchor and positive images as:
\begin{equation}
L_\text{T}(m_1,m_2) = \text{ReLU}(D(x^{m_1}_\text{anc}, x^{m_2}_\text{pos}) - D(x^{m_1}_\text{anc}, x^{m_2}_\text{neg}) + \epsilon),
\end{equation}
where $\epsilon$ is the margin. We then define our cross-domain hard example triplet loss as the sum of the within and between domain triplets:
\begin{equation}
    L_\text{Triplet} = L_\text{T}(A,A)+L_\text{T}(B,B)+L_\text{T}(A,B)+L_\text{T}(B,A).
\end{equation}

\paranospace{Cross Entropy Loss}
The triplet loss operates in the feature embedding space, with reference only to the feature embeddings of the images within the current batch. This means that these losses do not have access to any information that is shared between batches. For this reason, we add a classification head for the classes in $C_\text{train}$ and use a supervised classification cross-entropy loss for both domains. We find that this improves zero-shot performance. Note that the classification head is discarded at test time and only the feature embedding space used for zero-shot retrieval.

\subsection{Iterative Cluster-free Re-ranking (ICFRR)}

At test time, gallery images are ranked based on their similarity to a query image. This is done using Euclidean distance to measure dissimilarity in the image feature embedding space. However, since cross-domain feature comparison is very difficult, we propose a novel re-ranking scheme which enables correct class gallery images, that are initially low-ranked, to be moved higher up the ranked list based on gallery-gallery comparisons. Note that the classes of the query and gallery images remain unknown and we perform re-ranking independently for each query image.

Our approach uses an iteratively-updated similarity function, $s^{(t)}(x_Q^A,x_i^B)$. This provides a similarity score at iteration $t$ between the query image, $x_Q^A$, and each gallery image, $x_i^B$. Gallery images can be ranked at any iteration by sorting them by descending similarity score. 

We update the similarity score for each gallery image as:
\begin{equation}
    s^{(t+1)}(x_Q^A,x_i^B) := s^{(t)}(x_Q^A,x_i^B) + \beta\Delta_i^{(t)},\label{eqn:scoreupdate}
\end{equation}
where $\Delta_i^{(t)}\in\mathbb{R}_{\geq0}$ is the non-negative update to the similarity score (described below) and $\beta$ weights the influence of the update (and ultimately how re-ranking deviates from the original similarities). When the similarities are updated, query-gallery ranks change accordingly. We design the similarity score update to encourage correct matches to move up the ranked list.
We initialise our scores using negated query-gallery Euclidean distance:
\begin{equation}
    s^{(0)}(x^A_Q,x^B_i):=-D(x^A_Q,x^B_i).
\end{equation}
Hence, with zero iterations our procedure reduces to simply ranking by the original query-gallery distances.

\para{Gallery-gallery rank pre-computation} We begin by pre-computing all gallery-gallery distances using Euclidean distance in feature embedding space, $D(x^B_i,x^B_j)$, $\forall x^B_i,x^B_j\in\mathcal{G}$, and, for every gallery image, rank every other gallery image in ascending order of distance. We store these ranks in $r_{i,j}$ which gives the ranked position of gallery image $j$ against gallery image $i$ (i.e.~if all gallery images were sorted in increasing order of distance against image $i$ then $r_{i,j}$ tells us at what position in this sorted list gallery image $j$ would appear). In order to exclude self-matches from the ranked lists, we set $r_{i,i}=\infty$, $\forall i\in\{1,\dots,G\}$. These distances and ranks need only be computed once at very low cost ($O(G^2)$ for pairwise distances and $O(G^2\log G)$ to sort lists of distances into ranks with usually small $G$). This is negligible compared to the cost of computing the feature embeddings themselves and considerably cheaper than performing clustering of the gallery as in \cite{wang2019sketch,wang2019enhancing}. If an additional gallery image is added, we simply need to compute $G$ distances and insert the new element into the ranked lists. These fixed gallery-gallery ranks are subsequently used in our score update procedure. 

\begin{figure*}[!t]
    \centering
    \includegraphics[width=0.9\textwidth]{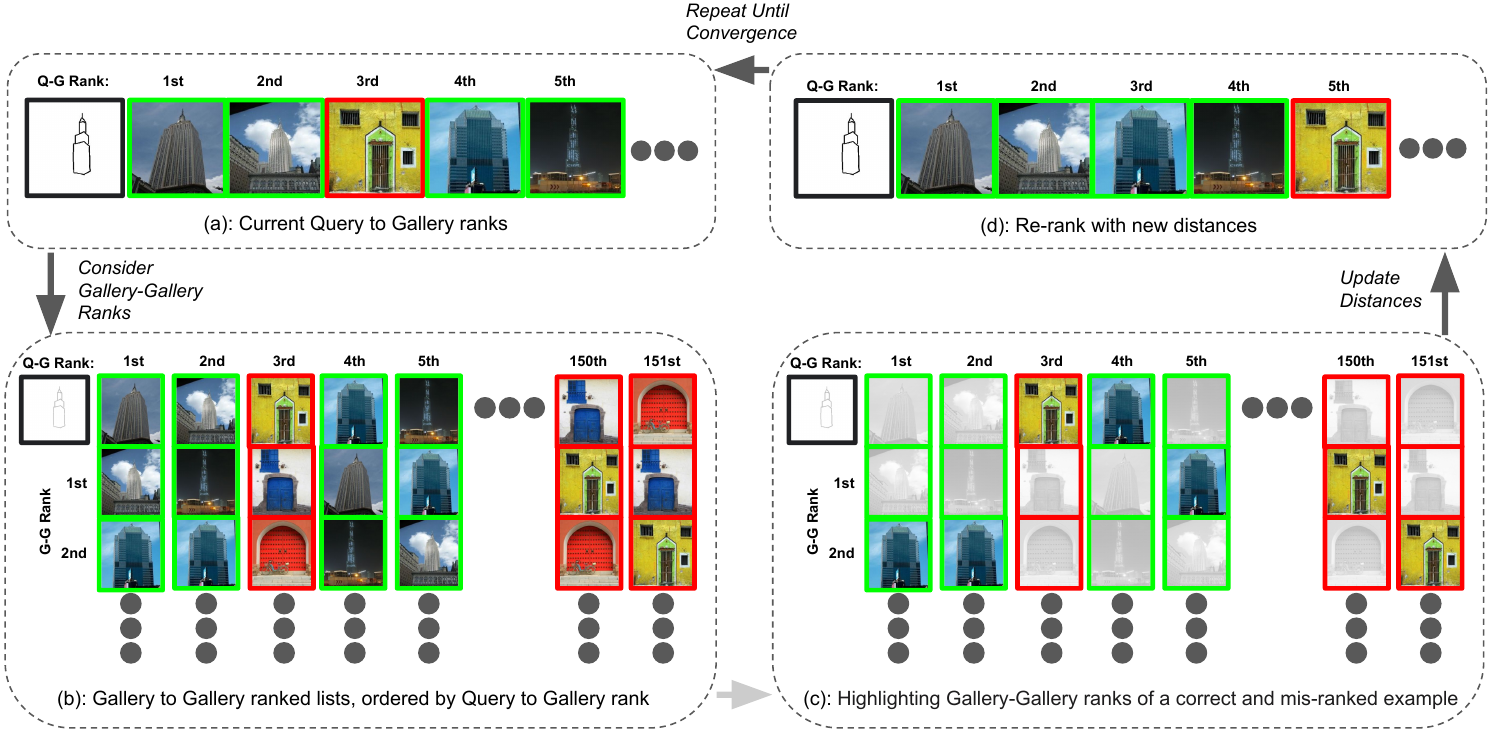}\hfill
    \caption{Overview of proposed iterative re-ranking approach. Initial query-gallery ranks (a) incorporate information from gallery-gallery ranks (b). Gallery images ranked highly against gallery images with high query-gallery rank (c) are rewarded and move up the ranked list when scores are updated (d).}\label{fig:reranking_combined_im}
\end{figure*}

\para{Rank-based similarity score update} The key to our approach is the update, $\Delta_i^{(t)}$, applied to the similarity score of each gallery image at each iteration. The idea is to reward gallery images that are ranked highly against gallery images that are themselves currently ranked highly against the query. Importantly, the similarity score update depends upon only rank information, not raw distances. The use of ranks provides robustness against extreme scores and means that we do not need to consider the distribution of distances or whether they must be normalised.

Using the similarity scores at iteration $t$, we can rank all gallery images against the query. We define $\mathcal{I}_p^{(t)}\in\{1,\dots,G\}$ as the index of the gallery image appearing at position $p$ in this ranked list, i.e.~$\mathcal{I}_p^{(t)}$ returns an integer in the range $1\dots G$ which is the index of a gallery image.

We consider only the current $K_q$ highest ranked gallery images against the query. Only gallery images currently ranked in the top $K_q$ positions against the query image can contribute to an increase in the similarity score of other gallery images against the query. Beyond this point, we assume there is little meaning in the rank order - these images are all considered distant from the query.

We use gallery-gallery ranks to update the similarity scores. However, we turn these ranks into scores:
\begin{equation}
    \alpha[r] = 
    \begin{cases}
    1-\frac{r-1}{G-1} & \text{if } r\leq K_g  \\
    0.0 & \text{otherwise}
    \end{cases}.
\end{equation}
This assigns a value of 1 to the highest ranked image. The score then reduces linearly with rank until position $K_g$ after which the score is 0. We only reward being ranked in the top $K_g$ positions in the gallery-gallery ranks. Again, the assumption is that beyond some point the rank order does not provide useful information.
We are now ready to define our similarity score update for the $i$th gallery image as:
\begin{equation}
    \Delta_i^{(t)} = \frac{1}{K_q}\sum_{p=1}^{K_q} \alpha[r_{\mathcal{I}_p^{(t)},i}].
\end{equation}
Note that this is a mean average over the currently $K_q$ best ranked gallery images. For each of these gallery images ($\mathcal{I}_p^{(t)}$, $p\in\{1,\dots,K_q\}$), we ask where gallery image $i$ is ranked against it ($r_{\mathcal{I}_p^{(t)},i}$) and convert this rank to a score using the $\alpha$ function.

The values of $K_q$, $K_g$ and $\beta$ become hyperparameters of our method, where the optimal choice for $K_q$ and $K_g$ depends on the size of the gallery and the typical number of images in each class. A good rule of thumb is to set $K_q$ and $K_g$ at around 0.5 times the number of instances you believe there to be of the positive class in the gallery set and $\beta$ to 0.5. We state the values used for each dataset within our Implementation Details.

\para{Iterative re-ranking} We repeatedly apply the similarity score update in \eqref{eqn:scoreupdate} until convergence (i.e.~ranks no longer change). We illustrate the behaviour of the re-ranking process in Figure \ref{fig:reranking_combined_im}. Here, the initial query-gallery ranking (a) places an incorrect match high in the list (marked in red). For every gallery image, we now consider their gallery-gallery ranks (b). In (c) we highlight the positions of two highly ranked examples in the other gallery ranked lists. Item 3 is an incorrect match. This is implied by the fact that it appears high in the ranked list of gallery images that are themselves ranked low against the query (items 150 and 151). Item 4 however is a correct match and this is implied by appearing high in the ranked list of gallery images that are themselves ranked highly against the query (items 1, 2 and 5). This will lead to a reward in the similairty score for item 4. Once the distances have been updated, we re-rank (d) and the correct matches move above the incorrect match which is pushed further down the query-gallery ranked list.

\begin{table*}[!t]
\resizebox{0.94\textwidth}{!}{
\begin{tabular}{lccccccccccccc}
\toprule
\multirow{2}{*}{Method} &
  \multicolumn{3}{c}{TU-Berlin Ext.} & &
  \multicolumn{3}{c}{Sketchy Ext.} & &
  \multicolumn{4}{c}{Sketchy Ext. \cite{yelamarthi2018zero} Split} \\
  \cmidrule{2-4} \cmidrule{6-8} \cmidrule{10-13}
 &
   
  \multicolumn{1}{c}{mAP@all} &
  \multicolumn{1}{c}{Prec@100} &
  \multicolumn{1}{c}{Prec@200} & &
  \multicolumn{1}{c}{mAP@all} &
  \multicolumn{1}{c}{Prec@100} &
  \multicolumn{1}{c}{Prec@200} & &
  \multicolumn{1}{c}{mAP@all} &
  \multicolumn{1}{c}{mAP@200} &
  \multicolumn{1}{c}{Prec@100} &
  \multicolumn{1}{c}{Prec@200}  \\ \hline \\[-8pt]
SBTKNet \cite{yelamarthi2018zero}         & 0.480  & 0.600  &  / &    & 0.550  & 0.690  & /    & & / & 0.500  & / & 0.590  \\
SAKE \cite{liu2019semantic}             & 0.475 & 0.599 & /   &  & 0.547 & 0.692 & /    & & / & 0.356 & / & 0.477  \\ 
TVT\textsuperscript{\textdagger} \cite{tian2022tvt}             & 0.484 & 0.662 & /   &  & 0.648 & 0.796 & /    & & / & 0.531 & / & 0.618 \\
NAVE \cite{wang2021norm}            & 0.493 & 0.607 & /    & & 0.613 & 0.725 & /    & & / & /     & / & /     \\
ACNet \cite{ren2023acnet}           & 0.577 & 0.658 & /    & & /     & /     & /    & & 0.559 & 0.517 & 0.643 & 0.608  \\
doodle2search \cite{dey2019doodle}  & 0.109 & /     & 0.121 & & 0.369 & /     & 0.370 & & / & /     & / & /      \\
RPKD \cite{tian2021relationship}           & 0.486 & 0.612 & /    & & 0.613 & 0.723 & /    & & / & 0.502 & / & 0.598 \\
PSKD\textsuperscript{\textdagger} \cite{wang2022prototype}           & 0.502 & 0.662 & /    & & 0.688 & 0.786 & /    & & / & 0.560 & / & 0.645 \\
ABDG \cite{tian2023zero}  & 0.489 & 0.686     & / & & 0.689 & 0.829  & / & & / & 0.567     & / & 0.649      \\
B-CLIP\textsuperscript{\textdagger} \cite{sain2023clip}         & 0.651 & 0.732 & / & & / & / & / & & / & 0.723     & / & 0.725  \\
MA-SBIR\textsuperscript{\textdagger, $\ddagger$} \cite{lyou2024modality}         & 0.705 & \textbf{0.777} & / & & / & / & / & & / & 0.691     & / & 0.755  \\
[2pt]
\hline \\[-8pt]
ViT-x\textsuperscript{\textdagger}                    & 0.695     & 0.720     & 0.718    & & 0.777     & 0.810     & 0.806    & & 0.682  & 0.747 &     0.737  & 0.722    \\
ViT-x\textsuperscript{\textdagger} + SuperGlobal \cite{shao2023global}                   & 0.707     & 0.611     & 0.607   &  & 0.778    & 0.810     & 0.806    & & 0.682 & 0.783   & 0.771 & 0.767   \\
ViT-x\textsuperscript{\textdagger} + AQE \cite{chum2007total}  & 0.695   & 0.734   & 0.739   &  &  0.777   &  0.835   &  0.834  & & 0.682  & 0.787 & 0.772  & 0.767    \\
ViT-x\textsuperscript{\textdagger} + DBA \cite{gordo2017end}  & 0.695  & 0.731  & 0.732   &  &  0.777   & 0.825  & 0.825   & & 0.682  & 0.784  & 0.772  & 0.771    \\
ViT-x\textsuperscript{\textdagger} + ECN \cite{sarfraz2018pose}  & 0.746  & 0.733  & 0.736   &  & 0.811    & 0.811     & 0.817    & & 0.731 & 0.746 & 0.733 & 0.741   \\

ViT-x\textsuperscript{\textdagger} + EDRM \cite{iscen2017efficient}  & 0.695  & 0.741  & 0.742   &  & 0.777    & 0.836     & 0.837    & & 0.695 & 0.731 & 0.720 & 0.718   \\
ViT-x\textsuperscript{\textdagger} + CSA \cite{ouyang2021contextual}   & 0.695  & 0.682  & 0.685   &&  0.777   & 0.789  & 0.801   & & 0.682 & 0.775 & 0.725 & 0.721   \\
\textbf{ViT-x\textsuperscript{\textdagger} + ICFRR }    & \textbf{0.754}     & 0.749     & 
\textbf{0.748}  &  & \textbf{0.836}    & \textbf{0.844}     & \textbf{0.845}    & & \textbf{0.742} & \textbf{0.790}     & \textbf{0.787} & \textbf{0.776}   \\
\bottomrule
\end{tabular}
}

\caption{A performance evaluation of our model against SOTA and common baseline approaches within ZS-SBIR and re-ranking using TU-Berlin \cite{eitz2012humans} and Sketchy \cite{liu2017deep, yelamarthi2018zero} datasets. We use both common test-set splits suggested for Sketchy. ``/'' represents metrics not reported in original published papers. ``*'' represents when we are not certain of which Sketchy dataset split the paper used. ``\textsuperscript{\textdagger}'' represents approaches that use a form of transformer network, all others use CNN based architecture. ``\textsuperscript{$\ddagger$}'' represents an approach that also inputs a text caption.}
\label{tab:baseline_comparisons}\vspace{-0.3cm}
\end{table*}

\begin{table}[!t]
\centering
\resizebox{1.0\columnwidth}{!}{
\begin{tabular}{lcccccccccccccccccc}
\toprule
\multirow{2}{*}{Method} &
  \multicolumn{4}{c}{QuickDraw Ext.} \\
  \cmidrule{2-5} 
 &
  \multicolumn{1}{c}{mAP@all} &
  \multicolumn{1}{c}{mAP@200} &
  \multicolumn{1}{c}{Prec@100} &
  \multicolumn{1}{c}{Prec@200} \\ \hline \\[-8pt]
doodle2search \cite{dey2019doodle}  & 0.075 & 0.090 & / & 0.067     \\
RPKD \cite{tian2021relationship}          &  0.143 & 0.128 & 0.230 & 0.218 \\
PSKD\textsuperscript{\textdagger} \cite{wang2022prototype}    & 0.150 & 0.199 & 0.297 & 0.298 \\
B-CLIP\textsuperscript{\textdagger} \cite{sain2023clip}  & 0.202 & / & / & 0.388  \\
MA-SBIR\textsuperscript{\textdagger, $\ddagger$} \cite{lyou2024modality}  & \textbf{0.327} & / & / & \textbf{0.425}  \\[2pt]
\hline \\[-8pt]
ViT-x\textsuperscript{\textdagger} &  0.204 & 0.306 & 0.279 & 0.272   \\
\textbf{ViT-x\textsuperscript{\textdagger} + ICFRR }  & 0.270 & \textbf{0.330} & \textbf{0.328} & 0.328 \\
\bottomrule
\end{tabular}
}

\caption{A performance evaluation of our model against SOTA and standard baseline approaches within ZS-SBIR using the QuickDraw\cite{dey2019doodle} dataset. ``/'' represents metrics not reported in original published papers. ``\textsuperscript{\textdagger}'' represents approaches that use a form of transformer network, all others use CNN based architecture. ``\textsuperscript{$\ddagger$}'' represents an approach that also inputs a text caption.}
\label{tab:quickdraw_comparisons}\vspace{-0.5cm}
\end{table}

\section{Experiments}

\paranospace{Datasets}
We evaluated our method on Sketchy Extended, TU-Berlin Extended and QuickDraw Extended, three commonly used datasets for the task of ZS-SBIR.
The original Sketchy dataset \cite{sangkloy2016sketchy} contained 12,500 photos and 75,479 sketches covering 125 different classes. \cite{liu2017deep} extended this dataset, adding another 60,502 photos bringing their total to 73,002. \cite{shen2018zero} randomly selected 25 classes to set as a test set, however as \cite{yelamarthi2018zero} noted, some of these classes have crossover with classes within ImageNet \cite{deng2009imagenet} and since it is common practice for models within SBIR to be initialised with weights pretrained on ImageNet, it did not match the zero-shot assumption. Hence, they propose a test set of 21 classes, carefully selected to not match image classes within ImageNet. %
We use both of these test setups.
TU-Berlin \cite{eitz2012humans} contains 204,489 photos and 20,000 sketches covering 250 different classes. Following, \cite{shen2018zero}, 30 classes are selected to be a test set. 
QuickDraw \cite{dey2019doodle} is created as a subset of the Google Quick, Draw! \cite{jong2016quic} data. It contains  330,000 sketches and 204,000 photos and covering 110 categories, 30 categories are used for a test set.

\para{Metrics}
In accordance with most state-of-the-art baseline implementations we will report both the mean Average Precision at k (mAP@k) and Precision at k (Prec@k) results. Due to the longevity of this topic of research, rank thresholds for which to measure the mAP and Prec have changed. We try to report all variations of k that appear in other work to give a consistent evaluation.

\para{Implementation Details}
We initialised our ViT model with pretrained ImageNet weights and used patchsizes of $16\times 16$ and feature dimensions of 768.
Adam \cite{kingma2014adam} optimiser was used with a weight decay of 5e-4. All pretrained layers are set to have a learning rate of 1e-5 while additional layers are initialised with a learning rate of 1e-4. 
${\lambda_\text{Triplet}}={\lambda_\text{CE}} = 1.0$
We trained for 80 epochs with a batchsize of 80. 
We use $\beta=0.5$ in all experiments, $K_q=K_g=512$ for TU-Berlin and $K_q=K_g=256$ for Sketchy and $K_q=100, K_g=125$ for the Sketchy \cite{yelamarthi2018zero} split. This difference is due to the reduced model accuracy in the \cite{yelamarthi2018zero} split, caused by the fully zero-shot nature of the test classes, leading to lower confidence in both query-gallery ranks and gallery-gallery ranks. The scale difference of $K_q$ and $K_g$ on TU-Berlin is due to the increased number of images present in the gallery dataset. 
Following \cite{zhai2019defense} we implement a class balancing dataloader using a PK sampling strategy, wherein each iteration involves randomly sampling a mini-batch of P classes. From each class and domain, K instances of images are sampled. As we focus on two domains in our retrieval tasks, each batch size is determined as 2PK, accommodating images from both domains. Having multiple instances of each class enhances the triplet loss learning process by providing a richer and more diverse set of positive samples. Moreover, having a range of unique classes in a batch ensures there are enough negative examples for the triplet loss to mine hard examples from, while the cross-entropy loss benefits from class diversity. We use P=5 and K=8 for our PK values.

\para{Benchmark Comparisons}
We report the ZS-SBIR performance of our approach against both state of the art and common baseline implementations in Table \ref{tab:baseline_comparisons} and Table \ref{tab:quickdraw_comparisons}. 
Our approach demonstrates superior performance across almost all baseline metrics, with the exception being the Prec@100 score for the TU-Berlin dataset. 
Improvements are also observed within our evaluation of the QuickDraw dataset, where we achieve state-of-the-art performance on mAP@200 and Prec@100 metrics but fall behind on the other two, with \cite{lyou2024modality} reporting a improved result.  
Also presented in Table \ref{tab:baseline_comparisons} are comparisons with other re-ranking approaches, encompassing both classical and contemporary methods. These include diffusion \cite{iscen2017efficient}, learned re-ranking \cite{ouyang2021contextual}, query expansion \cite{chum2007total}, feature augmentation \cite{gordo2017end} and k-nearest neighbours \cite{sarfraz2018pose, shao2023global}. Our re-ranking method consistently outperforms all these methods across all evaluated datasets.
As these methods did not report results on these datasets, we used their codebases to generate results, sweeping through parameter values to find a combination that reports the best results. Within the supplementary material we state experimental details as well as illustrate a comparison of the inference times across all the re-ranking methods we evaluated, including our proposed ICFRR method. Our results demonstrate that ICFRR achieves superior speed performance when run on a single iteration and maintains highly competitive performance even with multiple iterations compared to the single-iteration methods.

\begin{figure}[!t]
\centering
    \includegraphics[width=0.9\linewidth, trim=0cm 6.8cm 0cm 0cm, clip]{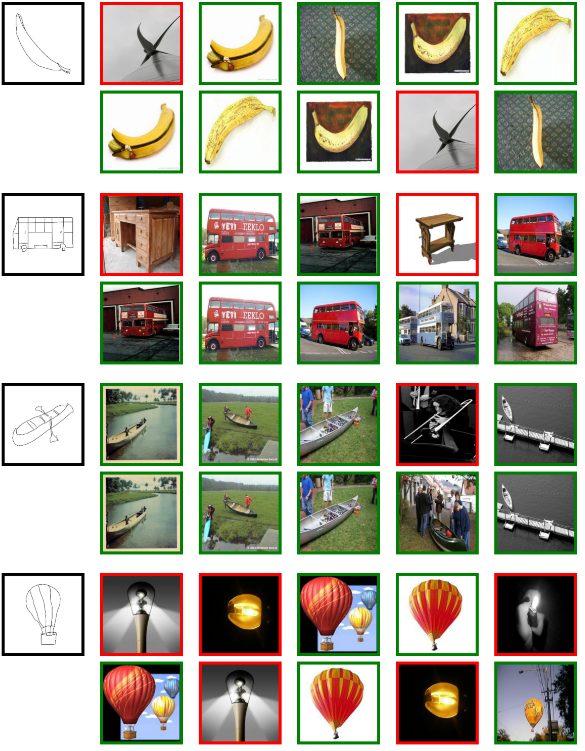}
    \caption{Results of re-ranking. For each query sketch the top row relates to the original output while the row below is the re-ranked output. Correct matches are shown in green, incorrect in red.}\label{fig:reranking examples_top_5}\vspace{-0.6cm}
\end{figure}

\para{Different Image Domains}
Our method is agnostic to the domains of the query and gallery images. Although most benchmarks for zero-shot cross-domain retrieval are for sketches and photographs, we can train on any pair of domains. Benchmarks for other pairs of domains do not exist so we have used an existing dataset for another task and use it for zero-shot cartoon-based photo retrieval. We chose the cartoon domain as it is different to sketch (contains shading and colouring) but still semantically very different to photos. Office-Home \cite{venkateswara2017deep} includes photographic and cartoon style images, so was chosen for this experiment. We use our re-ranking approach to confirm that our method is not only applicable to sketches and photographs but can be used to get high retrieval success from other domains. For this experiment we split the 65 classes into 55 classes for train and 10 classes for test, doing our upmost to ensure test classes were as far removed from ImageNet classifications as possible. Due to this being a small dataset there were only around 70 images per class, hence we report results on mAP@all and Prec@25. Presented in Table \ref{tab:office_home_ablation_comparisons}, we show increases in all metrics over both domain splits. The most notable enhancement is observed in the Art - Product domain, with an approximately 11\% increase in mAP@all. Additionally, we observe substantial improvements in the more semantically dissimilar domain pair of Clipart - Real World, with an approximate increase of 7\% in mAP@all.

\begin{table}[!t]
\centering
\begin{minipage}[b]{\columnwidth}
    \centering
    \resizebox{\columnwidth}{!}{
\begin{tabular}{@{}cccccccccc@{}}
\toprule
         & \multicolumn{2}{c}{SAKE \cite{liu2019semantic}} & 
         \multicolumn{2}{c}{ACNet \cite{ren2023acnet}} &
         \multicolumn{2}{c}{B-CLIP \cite{sain2023clip}} & \\
         & mAP@all & Prec@100 & mAP@all  & Prec@100 & mAP@all  & Prec@100
         \\ \cmidrule{1-7} 
Baseline & 0.470 & 0.621 & 0.559 & 0.643 & 0.691 & 0.754     \\
+ ICFRR & \textbf{0.546} & \textbf{0.672} & \textbf{0.644} & \textbf{0.676} & \textbf{0.758} & \textbf{0.793}   \\
\bottomrule
\end{tabular}
}\vspace{-0.2cm}
\caption{Applying our ICFRR to features from different backbones on Sketchy Ext.~\cite{yelamarthi2018zero} Split.}\vspace{0.2cm}
\label{tab:other_backbone_ablation_comparisons}
\end{minipage}
\hfill
\begin{minipage}[b]{0.47\textwidth}
    \centering
    \resizebox{0.8\columnwidth}{!}{
\begin{tabular}{@{}ccccccc@{}}
\toprule
         & \multicolumn{2}{c}{Clipart - Real World} & \multicolumn{2}{c}{Art - Product} \\ 
         & mAP@all & Prec@25 & mAP@all & Prec@25        \\ \cmidrule{1-5} 
ViT-x & 0.723 & 0.777 & 0.741 & 0.791         \\
+ ICFRR & \textbf{0.790} & \textbf{0.807} & \textbf{0.855} & \textbf{0.888}           \\
\bottomrule
\end{tabular}
}\vspace{-0.2cm}
\caption{ViT-x and our ICFRR on Office-Home \cite{venkateswara2017deep} domains.}\vspace{0.3cm}
\label{tab:office_home_ablation_comparisons}
\end{minipage}
\end{table}

\begin{table}[!t]
\centering
\vspace{-0.5cm}
\resizebox{\columnwidth}{!}{
\begin{tabular}{ccccccccc}
\hline
\multirow{2}{*}{CE Loss} &
\multirow{2}{*}{Triplet Loss} &
  \multirow{2}{*}{ICFRR} &
  \multicolumn{2}{c}{TU-Berlin Ext.} &
  \multicolumn{2}{c}{Sketchy Ext.} &
  \multicolumn{2}{c}{Sketchy Ext. \cite{yelamarthi2018zero} Split} \\  
  &   &   & mAP@all & Prec@100 & mAP@all & Prec@100 & mAP@all & Prec@100 \\ \cline{1-9} \\[-10pt]
N & Y & N & 0.483   & 0.561    & 0.464   & 0.574    & 0.483   & 0.558    \\
Y & N & N & 0.553   & 0.635    & 0.621   & 0.722    & 0.459   & 0.555    \\
Y & Y & N & 0.693   & 0.720    & 0.777   & 0.810    & 0.682   & 0.737    \\
Y & Y & Y & 0.754   & 0.749    & 0.836   & 0.844    & 0.742   & 0.787    \\ \hline
\end{tabular}}\vspace{-0.2cm}
\caption{Ablation study. CE Loss is Cross Entropy Loss, Triplet Loss is Cross Domain Hard Mining Triplet Loss and ICFRR is our Iterative Cluster-free Re-ranking.}\vspace{-0.3cm}
\label{tab:ablation_study}
\end{table}

\para{Ablation Study}
In Table \ref{tab:ablation_study}, we assess the impacts of the main components of our implementation. Building on the solid foundation created by a Cross Entropy and Cross-domain Hard-Mining Triplet loss, our Re-ranking approach gives a large increase of around 5-8\% on the benchmark datasets. This highlights our ICFRR's ability to bridge the semantic gap between sketch and images. 

\para{Re-ranking with Other Models} Our re-ranking method can be applied to features from any backbone. In Table \ref{tab:other_backbone_ablation_comparisons} we show results using features from the seminal SAKE \cite{liu2019semantic} and contemporary ACNet \cite{ren2023acnet} and B-CLIP \cite{sain2023clip} methods. Compared to their baseline retrieval results, our re-ranking always improves performance, by as much as 8.5\% in mAP@all.

\begin{figure}[!t]
    \centering
    \includegraphics[width=0.9\linewidth]{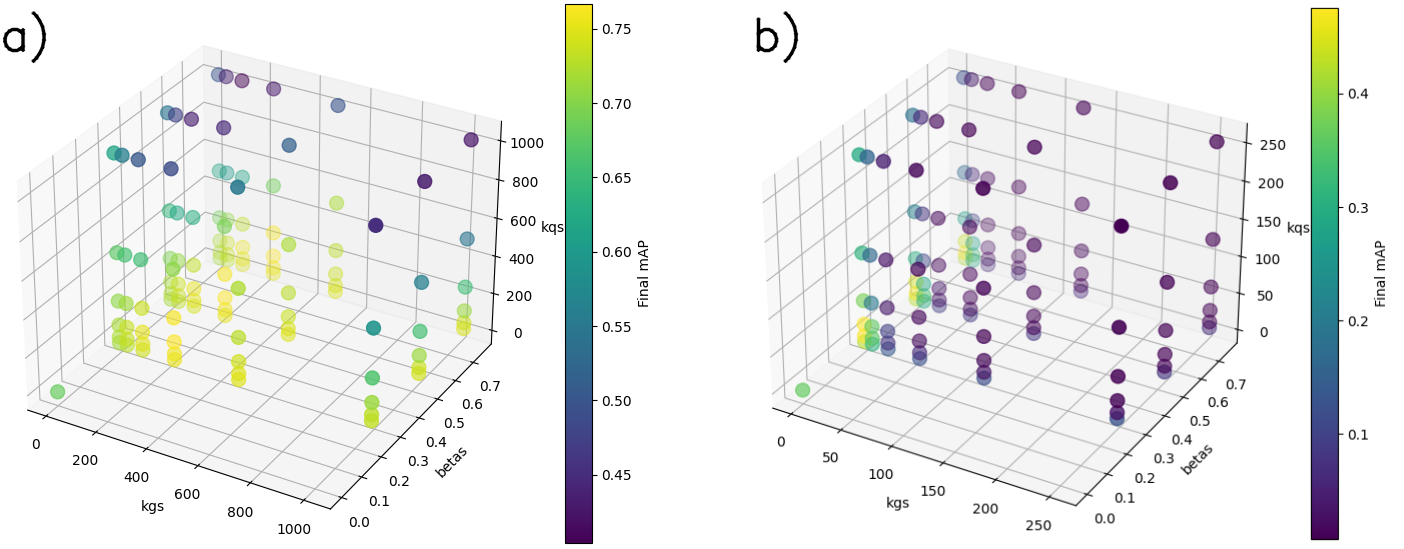}
    \caption{Varying $K_q$, $K_g$ and $\beta$ on Sketchy dataset \cite{yelamarthi2018zero}; \textit{a} shows full dataset, \textit{b} a single query class with 10 gallery examples.}
    \label{fig:parameter_sweep_comb}
\end{figure}

\para{Effect of $K_q$, $K_g$ and $\beta$ Hyperparameters}
In Fig.~\ref{fig:parameter_sweep_comb}a we present sweeps across the three hyperparameters along with the baseline non re-ranked value. This shows that our method is relatively insensitive to the precise values, consistently outperforming the baseline without re-ranking. However, extremely high values of $K_q$ and $K_g$ led to instability while very low values had negligible impact over the baseline. Minor effects were observed with $\beta$ changes.  
We also test the case when there is a query with very few matching instances in the gallery set. To achieve this we isolate one query class and reduce its true positives within the gallery set to only 10, compared to the typical 500-700 images per class. The result is shown in Fig.~\ref{fig:parameter_sweep_comb}b. With appropriate selection of $K_q$ and $K_g$, reflecting the low expected true positive rate, our ICFRR still improves upon baseline results.

\section{Conclusions}
We propose a novel test time method for retrieval tasks across different domains. We focus our evaluation on ZS-SBIR while emphasising the domain-agnostic versatility our approach allows for, demonstrated through our comprehensive ablation studies. By unifying a ViT model and zero-shot retrieval losses, we train a high-performing model to act as our baseline. During testing, we illustrate how our re-ranking approach exploits within-domain semantic relationships to enhance cross-domain ranking. Our approach achieves state-of-the-art performance against ZS-SBIR and re-ranking baselines across multiple datasets, surpassing many mAP@all metrics by a significant margin. Moreover, we show the model-agnostic nature of the re-ranking, enhancing the retrieval success of the SAKE and ACNet models. A promising extension would be to leverage information within model output patches to establish more specific feature space relationships between images, enabling a finer semantic understanding of the category representation.

{\small
    \bibliographystyle{ieeenat_fullname}
    \bibliography{refs}
}

\clearpage
\setcounter{page}{1}
\maketitlesupplementary

\section{Supplementary Material}

\subsection{Iterative Behaviour of Re-ranking}
\begin{figure*}[!t]
    \centering
    \centering
    \includegraphics[width=\textwidth]{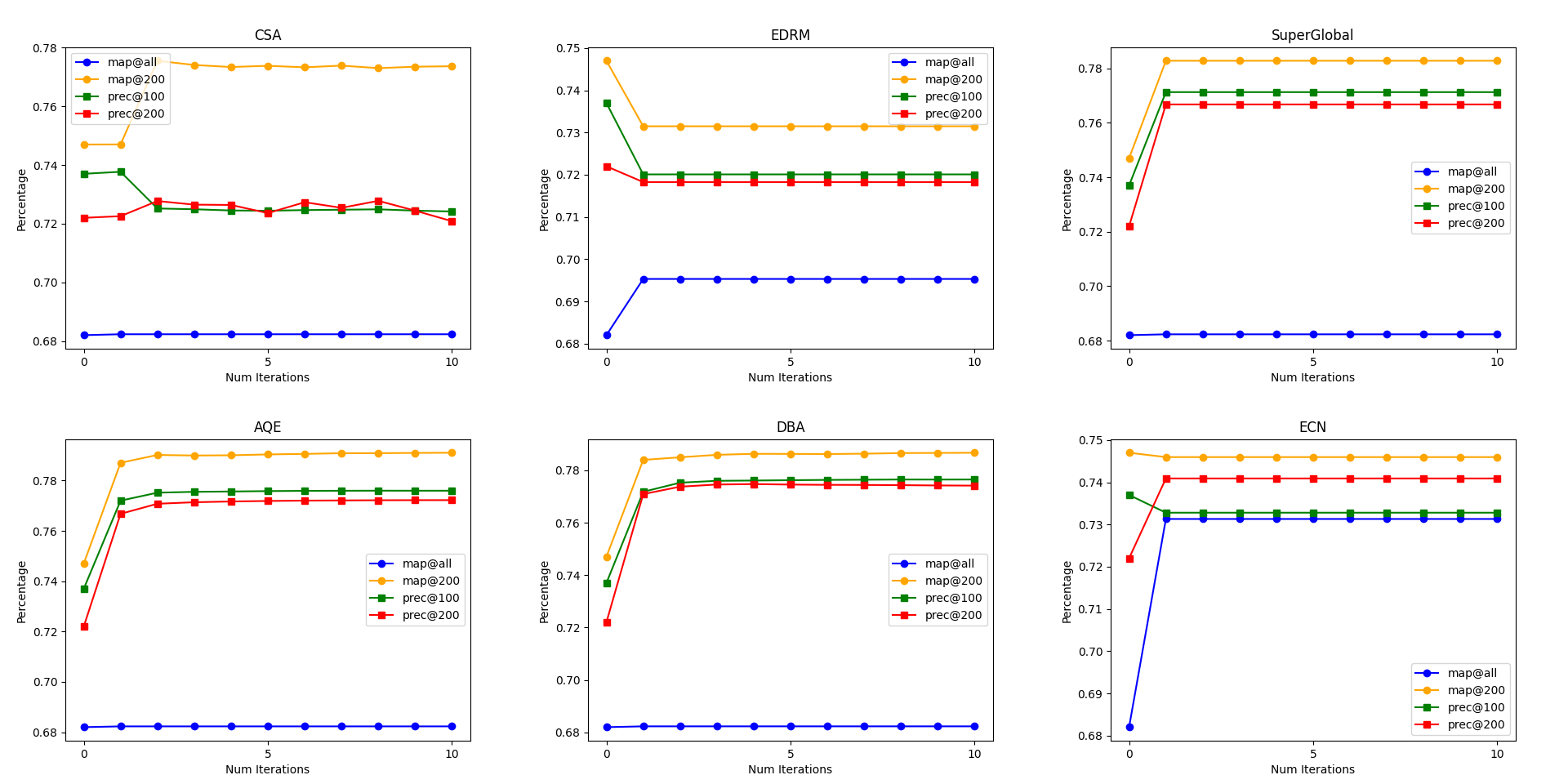}\hfill
    \caption{Iterative reranking does not improve other methods}\label{fig:iterative_doesnt_help_others}
\end{figure*}

As mentioned, our re-ranking method is an iterative approach; updated ranks are used to recompute the query-gallery distances which in turn update the ranks. 
Figure \ref{fig:sketchy_converging} portrays the iterative process of re-ranking. The performance metrics for the Sketchy dataset are plotted against the number of re-ranking iterations. The optimal iteration point for all metrics emerges at around iteration 10, exhibiting minimal variations thereafter. Although iteration 5 sees an improvement in the top 200 ranks, the mAP@all value has yet to stabilise. The decrease in top 200 precision, compared to iteration 5, is attributed to the progressively longer range connections in later iterations, which can cause some mistaken re-ranks. However, even at iteration 10 the mAP@all value is still a $\sim$4\% improvement compared to baseline rank results. We extend this ablation to include other re-ranking methods \cite{shao2023global, iscen2017efficient, ouyang2021contextual, gordo2017end, chum2007total, sarfraz2018pose} and showcase, in figure \ref{fig:iterative_doesnt_help_others}, that just running these methods for more than one iteration does not yield better results. We attribute this to the fact that our approach updates all gallery scores with each iteration, whereas other methods operate within a more restricted search space and thus fail to explore distant relationships effectively.

\subsection{Inference Speed}
\begin{figure}[!t]
    \centering
    \includegraphics[width=.7\columnwidth]{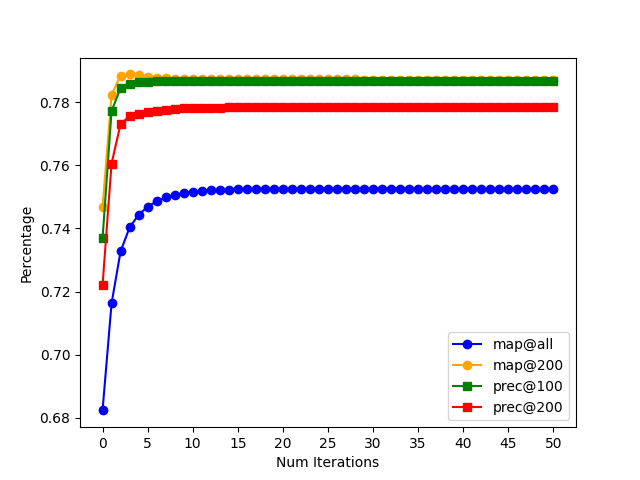}\hfill
    \caption{Convergence of re-ranking approach on the Sketchy \cite{yelamarthi2018zero} split dataset.}\label{fig:sketchy_converging}
\end{figure}

\begin{figure}[!t]
    \centering
    \includegraphics[width=\columnwidth]{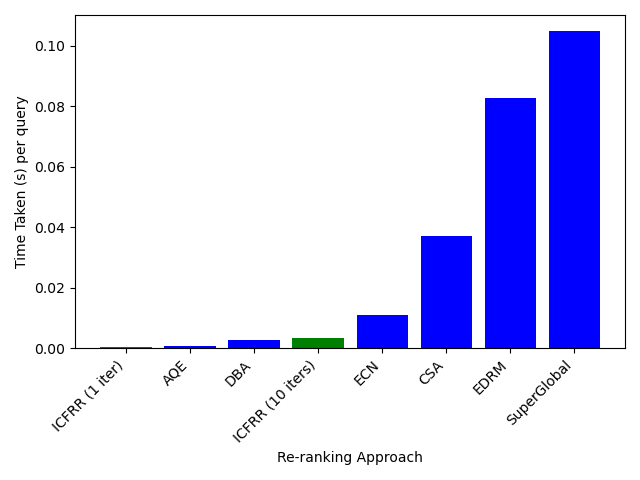}\hfill
    \caption{Inference time measured for re-ranking a single query against the Sketchy \cite{yelamarthi2018zero} split (average computed over all queries in the test set). We showcase speed for a single iteration of ICFRR and 10 iterations, which was used when getting benchmark results for this dataset as shown in Figure \ref{fig:sketchy_converging}. }\label{fig:inference_time}
\end{figure}

Figure \ref{fig:inference_time} displays the inference time of a single re-ranking iteration of a query image retrieval of the Sketchy \cite{yelamarthi2018zero} split dataset. Our ICFRR method demonstrates superior efficiency with a single iteration, requiring only 50\% of the time of the next fastest method (AQE \cite{chum2007total}) and just 0.33\% of the time compared to the slowest method (SuperGlobal \cite{shao2023global}). When our ICFRR is ran for multiple iterations, 10 in the case of this dataset (Figure {\ref{fig:sketchy_converging}}) we demonstrate that our method achieves speeds comparable to or faster than other re-ranking approaches which just use single iterations.
Please bear in mind that these findings were achieved utilising an NVIDIA 4080 GPU and 32GB of RAM. Speeds may fluctuate across diverse platforms and, naturally, when working with different datasets. Additionally, we acknowledge that our code implementation might not be optimised for maximum speed efficiency, and employing threading strategies could likely further enhance the speed.

\subsection{Benchmark Re-ranking Comparisons}
\begin{figure}[!t]
    \centering
    \includegraphics[width=\columnwidth]{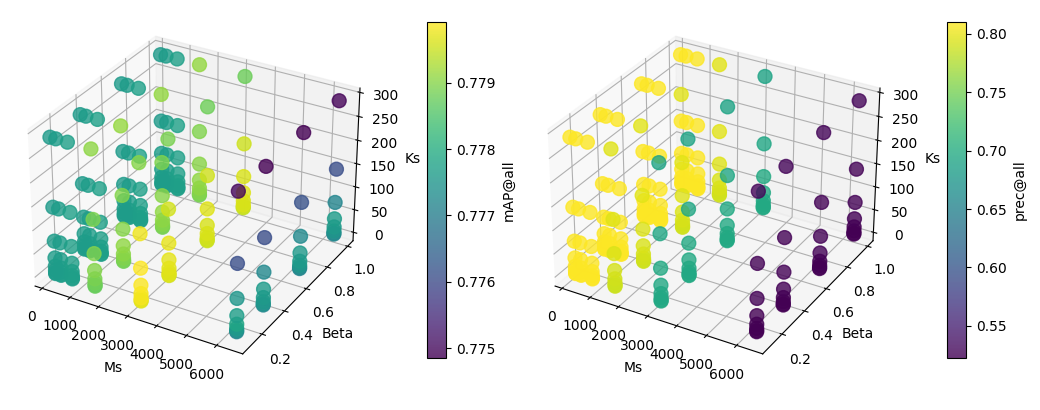}\hfill
    \caption{Parameter sweep of Super Global method on Sketchy dataset.}\label{fig:super_global_param_sweep}
\end{figure}

\begin{figure}[!t]
    \centering
    \includegraphics[width=\columnwidth]{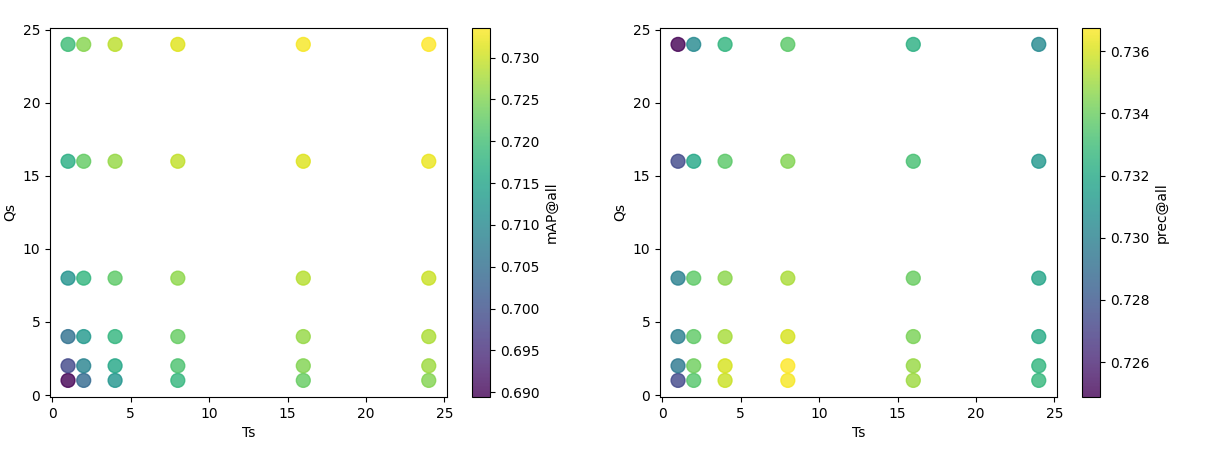}\hfill
    \caption{Parameter sweep of ECN method on Sketchy dataset with \cite{yelamarthi2018zero} split.}\label{fig:ecn_param_sweep}
\end{figure}

As mentioned in \textit{Benchmark Comparisons} we compared our ICFRR results against SuperGlobal \cite{shao2023global}, AQE \cite{chum2007total}, DBA \cite{gordo2017end}, ECN \cite{sarfraz2018pose}, EDRM \cite{iscen2017efficient} and CSA \cite{ouyang2021contextual} methods.
These methods have no reported results for any of the benchmark datasets used in this paper. Hence, we have had to run these method's code releases on these datasets to generate results. We wanted to be as fair as we possibly could to these implementations, so we ran parameter sweeps for each implementation to find the best possible combination of parameters to give these implementations the highest output results. ECN \cite{sarfraz2018pose} uses query-query distances directly, which as mentioned in this paper is not allowed for the ZS-SBIR paradigm, we therefore have removed any effect of query-query distances within the method. Figure \ref{fig:super_global_param_sweep} and Figure \ref{fig:ecn_param_sweep} show the parameter sweeps for the SuperGlobal and ECN methods respectively. To report the best benchmark we pick the parameters that provide the best midpoint for mAP@all and Prec@100 results.

\subsection{Examples of ZS-SBIR Results with Re-ranking}
\begin{figure*}[!t]
    \includegraphics[width=0.8\textwidth]{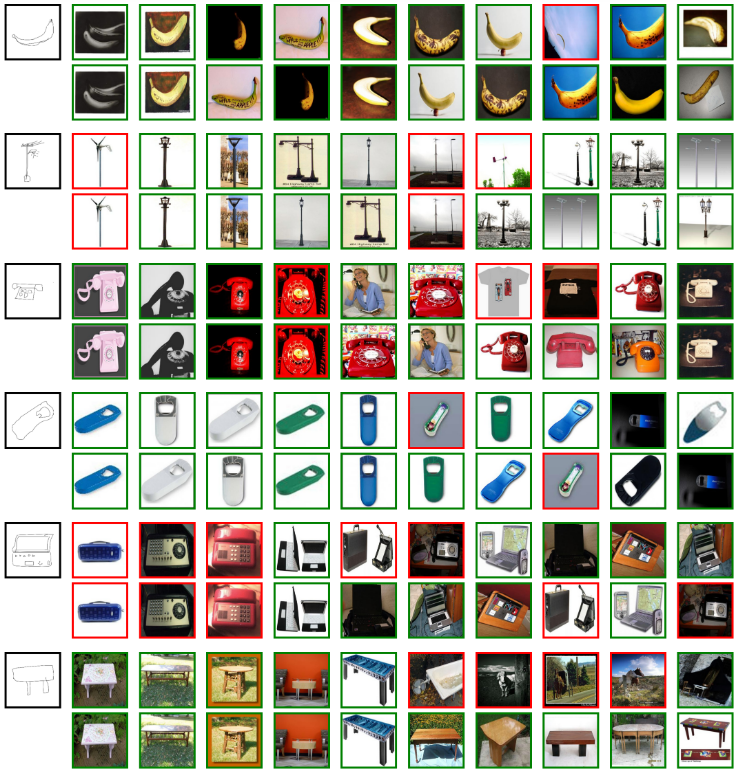}\hfill
    \includegraphics[width=0.8\textwidth]{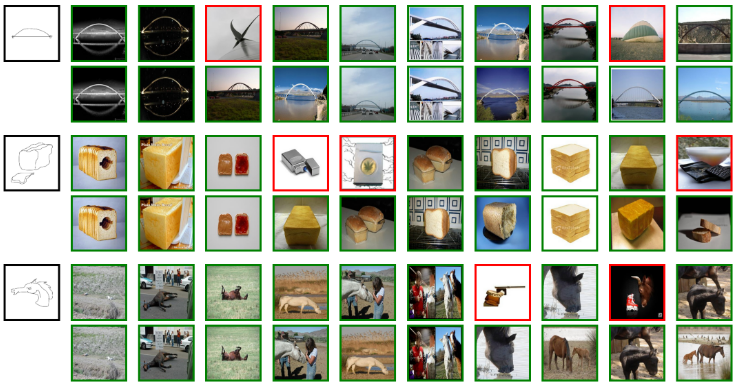}\hfill
    \caption{Extended examples of re-ranking improving ranking with TU Berlin dataset. For each query sketch the top row relates to the original output while the row below is the re-ranked output. Correct matches are shown in green, incorrect in red}.\label{fig:reranking examples_extended}
\end{figure*}

\begin{figure*}[!t]
    \includegraphics[width=\textwidth]{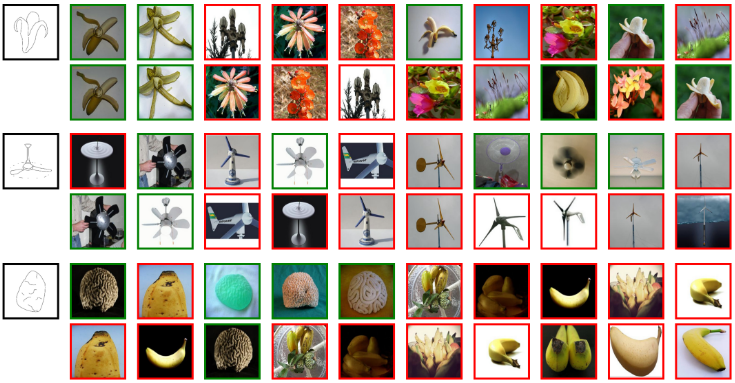}\hfill
    \caption{Examples of re-ranking harming the ranking with TU Berlin dataset. For each query sketch the top row relates to the original output while the row below is the re-ranked output. Correct matches are shown in green, incorrect in red.}\label{fig:reranking bad_examples_extended}
\end{figure*}

In Figures \ref{fig:reranking examples_extended} and \ref{fig:reranking bad_examples_extended} we showcase some further results of our re-ranked top-10 retrieval. Figure \ref{fig:reranking examples_extended} highlights some of the successes of our re-ranking approach on the TU-Berlin dataset. One particular example to draw attention to is the sketch of a horse (located at the bottom of the figure). The model has assessed this sketch as being very similar to the handgun, ranked in 7th place, for understandable reasons. The handgun shares a very similar shape to that of the sketched horse head, but very little similarity with the highly ranked photographs, so our re-ranking method successfully moves this incorrect image out of the top 10 rankings. 

Figure \ref{fig:reranking bad_examples_extended} shows failure cases of our re-ranking method where it actually harms retrieval performance of the model. Our re-ranking is less successful when only a small subset of the top results are correctly ranked, weighting the very top examples more highly in its re-ranking method. However, as the top example (banana sketch) shows, even when the top 2 are correct the method can still fail to improve the rankings. This could be due to these images not having many photo domain similarities with photos of the same class so the re-ranking fails to move any of the same class up the ranks and, in fact, moves some of the correct class examples down the ranked list.

\subsection{Further Examples of other Cross-domain Dataset's Results with Re-ranking}
\begin{figure*}[!t]
    \centering
    \includegraphics[width=\textwidth]{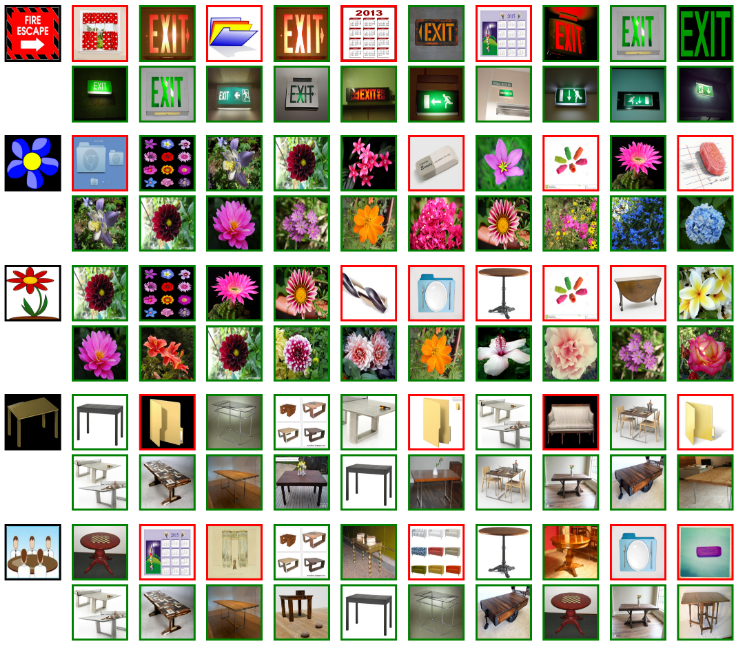}\hfill
    \caption{Extended examples of re-ranking improving ranking with Office-Home dataset. For each query sketch the top row relates to the original output while the row below is the re-ranked output. Correct matches are shown in green, incorrect in red}\label{fig:officehome_reranking examples_extended}
\end{figure*}

\begin{figure*}[!t]
    \centering
    \includegraphics[width=\textwidth]{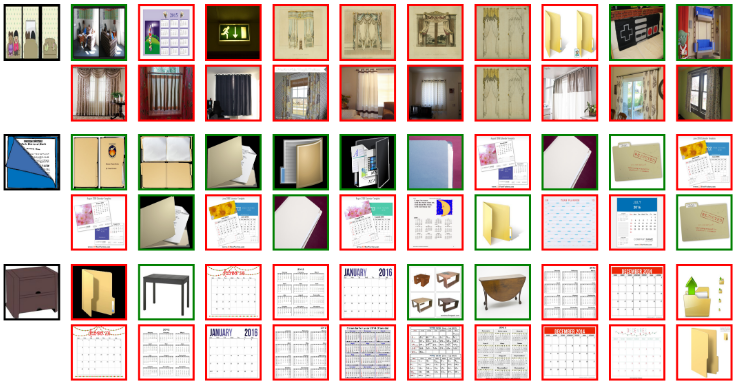}\hfill
    \caption{Examples of re-ranking harming the ranking with Office-Home dataset. For each query sketch the top row relates to the original output while the row below is the re-ranked output. Correct matches are shown in green, incorrect in red}\label{fig:officehome_reranking bad_examples_extended}
\end{figure*}

In Figure \ref{fig:officehome_reranking examples_extended} and Figure \ref{fig:officehome_reranking bad_examples_extended} we showcase some further results of our re-ranked top-10 retrieval on the Office Home dataset. Figure \ref{fig:officehome_reranking examples_extended} demonstrates how our ICFRR is capable of eliminating erroneous gallery images that lack similarity with any other top-10 images within the photographic domain. In the second example involving a clip-art blue "flower", we observe that erroneous values of the folder (which was original top ranked), and other classes, which shared some colour or shape similarities to the blue "flower" are moved below rank 10. Instead the "flower" photographs, which share a semantic similarity to the query blue "flower", are promoted higher in the rankings.

In contrast, Figure \ref{fig:officehome_reranking bad_examples_extended} highlights some failure cases of our ICFRR on the Office Home dataset. The first and third examples highlight situations in which the initial ranking, solely based on the model's feature similarities, yields poor-quality results, resulting in inaccurate top-ranked outcomes. In the second example, the initial ranking outperforms the ICFRR ranking. This discrepancy can be attributed to the high similarity between the "folder" and "calendar" classes in the photographic domain. During the iterative re-ranking, the connections between the "calendar" classes are likely stronger, resulting in their promotion up the ranks in comparison to the "folder" classes.

\begin{figure*}[!t]
\centering
    \includegraphics[width=\textwidth,clip=true,trim=0px 260px 0px 0px]{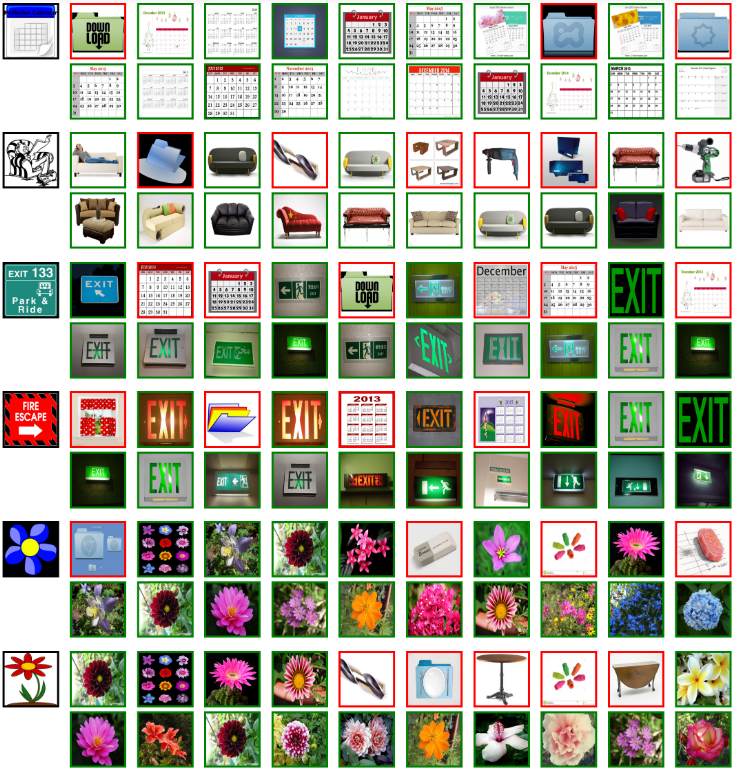} %
    \caption{Examples showing result of Office-Home re-ranking. For each query sketch the top row relates to the original output while the row below is the re-ranked output. Correct matches are shown in green, incorrect in red.}\vspace{-0.3cm}\label{fig:office home reranking examples} %
\end{figure*}

\end{document}